\documentclass[conference]{IEEEtran}
\IEEEoverridecommandlockouts
\usepackage{eso-pic}

\usepackage{cite}
\usepackage{amsmath,amssymb,amsfonts}
\usepackage{algorithm}
\usepackage{algpseudocode}
\usepackage{graphicx}
\usepackage{textcomp}
\usepackage{xcolor}
\usepackage{booktabs}
\usepackage{tabularx}
\usepackage{multirow}
\usepackage{placeins}
\usepackage{listings}
\usepackage{colortbl}
\definecolor{stagerow}{gray}{0.92}

\def\BibTeX{{\rm B\kern-.05em{\sc i\kern-.025em b}\kern-.08em
    T\kern-.1667em\lower.7ex\hbox{E}\kern-.125emX}}

\lstdefinelanguage{json}{
    basicstyle=\ttfamily\small,
    numbers=left,
    numberstyle=\tiny,
    stepnumber=1,
    numbersep=6pt,
    stringstyle=\color{blue},
    showstringspaces=false,
    morestring=[b]",
    morecomment=[l]{//},
    moredelim=[l][\color{magenta}]{:\ },
}

\renewcommand\IEEEkeywordsname{Keywords}

\begin{document}
\AddToShipoutPictureBG*{%
  \AtPageUpperLeft{%
    \put(\LenToUnit{0.5\paperwidth},\LenToUnit{-1.2cm}){%
      \makebox[0pt][c]{%
        \begin{minipage}{0.9\paperwidth}
          \centering \footnotesize
          Hieu Minh Duong, Todd Gary, Eugene Levin, and Long Nguyen. ``STAGE: Spatio-Temporal Augmented Graph Expansion for Weakly Supervised Cross-Tweet Event Causality Discovery'', accepted at IEEE MLBD 2026.
        \end{minipage}%
      }
    }
  }
  \AtPageLowerLeft{%
    \put(\LenToUnit{0.5\paperwidth},\LenToUnit{1.0cm}){%
      \makebox[0pt][c]{%
        \begin{minipage}{0.9\paperwidth}
          \centering \scriptsize
          \copyright 2026 IEEE. Personal use of this material is permitted. Permission from IEEE must be obtained for all other uses, in any current or future media, including reprinting/republishing this material for advertising or promotional purposes, creating new collective works, for resale or redistribution to servers or lists, or reuse of any copyrighted component of this work in other works.
        \end{minipage}%
      }
    }
  }
}

\title{STAGE: Spatio-Temporal Augmented Graph Expansion for Weakly Supervised Cross-Tweet Event Causality Discovery}

\author{%
\IEEEauthorblockN{Hieu Minh Duong\IEEEauthorrefmark{1},
Todd Gary\IEEEauthorrefmark{2},
Eugene Levin\IEEEauthorrefmark{2}, and
Long Nguyen\IEEEauthorrefmark{1}}
\IEEEauthorblockA{\IEEEauthorrefmark{1}University of Louisville, Louisville, Kentucky, USA\\
\{hieu.duong, l.nguyen\}@louisville.edu}
\IEEEauthorblockA{\IEEEauthorrefmark{2}Meharry Medical College, Nashville, Tennessee, USA\\
\{tpgary, elevin\}@mmc.edu}
}

\maketitle

\begin{abstract}
Causal relations in disaster-related social media often span independent posts, making cross-tweet causal discovery difficult to supervise and model at scale. We propose STAGE, a framework that separates scalable cross-tweet weak-supervision construction from spatio-temporal relational learning over event occurrences. Starting from local causal seeds, STAGE retrieves semantically related event occurrences across tweets and uses a causal scorer to construct weakly labeled cross-tweet pairs. These pairs supervise a multi-head Graph Attention Network over semantically enriched event representations and a spatio-temporal context graph. We evaluate STAGE on California Wildfire and Hurricane Harvey using human-gold cross-tweet test sets and 15 baselines. Compared with local-only supervision, the constructed cross-tweet supervision improves STAGE AUPRC by 0.340 on Wildfire and 0.107 on Hurricane. STAGE achieves in-domain AUROC of 0.885 on Wildfire and 0.710 on Hurricane. Cross-disaster experiments reveal asymmetric transfer, highlighting sensitivity to dataset-specific spatio-temporal shifts. These results demonstrate the value of scalable cross-tweet supervision combined with relational event modeling for causal learning across fragmented disaster reports.
\end{abstract}

\begin{IEEEkeywords}
Natural Language Processing, Causal Discovery, Spatio-Temporal Graphs, Weak Supervision
\end{IEEEkeywords}

\section{Introduction}

\begin{figure}
    \centering
    \includegraphics[width=\linewidth]{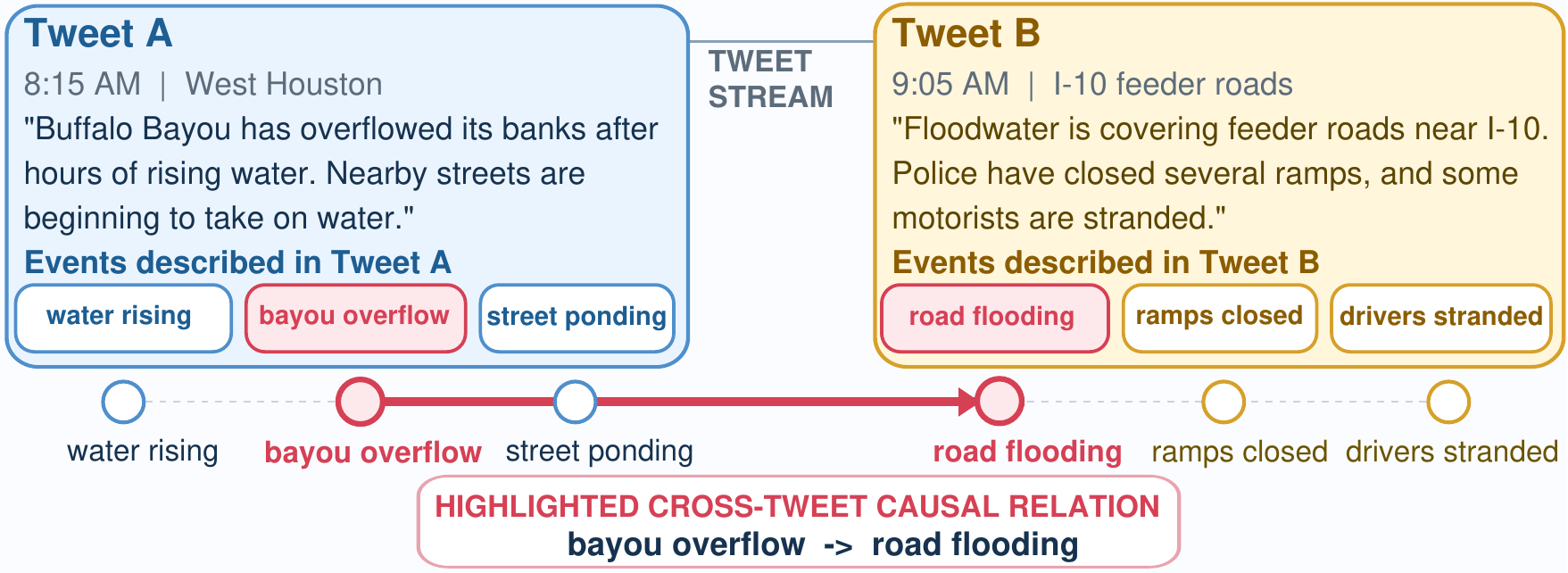}
    \caption{Illustrative example of cross-tweet causality. Each tweet describes multiple event instances, while the highlighted relation connects a specific event described in Tweet A to a specific event described in Tweet B.}
    \label{fig:example}
    \vspace{-4mm}
\end{figure}

Discovering causal relationships in social media presents a fundamental challenge, particularly for understanding how real-world events unfold and escalate during disasters. Traditional causal extraction methods are fundamentally limited for uncovering causal chains in social media streams because they operate on isolated text units. Sentence-level approaches rely on explicit linguistic cues that are rare in informal tweets. Neural and LLM-based models capture semantic associations but may conflate them with causation due to a lack of structural grounding \cite{anuyah2025}. Meanwhile, cross-document event linking methods such as Zhao et al.~\cite{zhao} rely on richer document-level context, while knowledge-augmented approaches may depend on retrieved evidence \cite{lewis2020rag} or structured event graphs \cite{roy}, making them difficult to apply directly to noisy, fragmented, and high-volume disaster streams. Figure~\ref{fig:example} illustrates this cross-tweet setting, where a causal relation connects specific event occurrences described in different tweets rather than the tweets as whole units.

The core challenge is therefore to move beyond single-text analysis and develop a method that can connect causal information distributed across posts while exploiting the spatio-temporal structure of an unfolding disaster. An effective solution should not rely solely on explicit textual markers, but should also exploit temporal, spatial, and semantic context when reasoning about relations between specific events. We introduce STAGE (Spatio-Temporal Augmented Graph Expansion), a framework that combines scalable cross-tweet weak supervision with spatio-temporal relational learning. STAGE constructs weakly supervised cross-tweet event pairs and models them through an event graph grounded in three complementary signals: (1) temporal information that captures temporal proximity and reporting context, (2) geographic information that captures spatial relationships among events, and (3) semantically enriched event representations that bridge lexical variation across posts.

In this work, we use causal discovery in the task-specific sense of identifying directed causal relations between real-world event instances described across social media posts. The prediction unit is an ordered pair of event occurrences, each representing a specific event instance described in a particular tweet. A cross-tweet pair therefore consists of two occurrences drawn from different tweets, and the prediction determines whether one causes the other and, if so, in which direction. This formulation enables event-level reasoning rather than treating entire tweets as causal units.

The framework separates cross-tweet supervision construction from spatio-temporal graph learning. Cross-tweet candidate pairs are first generated from semantic neighborhoods defined by event triggers and arguments, without temporal or spatial constraints, providing the basis for scalable weak supervision. STAGE then combines semantic, temporal, and spatial information, applies semantic expansion across related event descriptions, and constructs a graph whose contextual edges capture spatio-temporal relationships. A Graph Attention Network propagates information over this structure before predicting causal existence and direction for ordered event pairs. Thus, temporal and spatial information contribute to causal learning without determining which pairs enter the candidate pool. Our contributions are threefold:
\begin{itemize}
\item We propose STAGE, a framework for cross-tweet causal discovery that separates scalable cross-tweet supervision construction from spatio-temporal relational learning over event occurrences.

\item We introduce a scalable weak-supervision strategy for cross-tweet causal learning in which candidate event pairs are constructed from trigger-and-argument semantic neighborhoods and assigned weak causal labels by a fixed causal scorer, while human-gold pairs are withheld from model training and reserved for evaluation.

\item We instantiate the relational learner with a spatio-temporal event graph and multi-head GAT, and evaluate the framework under in-domain weak supervision, local-to-cross-tweet generalization, cross-disaster transfer, and component ablations.
\end{itemize}

\section{Related Work}
\subsection{Causal Discovery in Natural Language}

Early causal extraction relied on lexical and syntactic patterns \cite{khoo}, semantic parsing for causal questions \cite{girju2003}, and feature-based classifiers using dependency and lexical cues \cite{blanco2008}, with resources such as SemEval-2010 Task 8 supporting supervised relation learning \cite{semeval2010task8}. Later work incorporated external structure and data augmentation: LSIN uses induced descriptive and relational graphs \cite{cao2021}, while knowledge-guided augmentation \cite{zuo2021} and prompt-based auxiliary objectives \cite{shen2022} improve event causality identification. Furthermore, recent methods increasingly model richer event interactions. SemDI formulates event causality as semantic dependency inquiry \cite{semdi}; ICCL uses contrastive demonstrations \cite{iccl}; iLIF iteratively constructs and exploits an event causality graph \cite{ilif}; and GLM4ECI evaluates LLM-generated statements for directional event causality identification \cite{glm4eci}. More recently, Su et al.\ propose a suggest--verify--revise framework with structural refinement for document-level causality \cite{su-etal-2026-suggest}. These methods advance event causality identification, but primarily operate within sentences or coherent documents, whereas disaster tweets are short, noisy, and may distribute causal evidence across independent posts.

\subsection{Weak Supervision for Event and Relation Learning}

Weak supervision reduces annotation cost by deriving labels from indirect signals. Snorkel combines heuristic labeling functions through a probabilistic label model \cite{ratner2017}, while distant supervision aligns textual mentions with knowledge-base relations to construct large weakly labeled corpora \cite{mintz2009}. For event relations, Yao et al.\ derive weak temporal labels from ordering regularities \cite{yao2017weak}, while Mirroshandel and Ghassem-Sani infer temporal relations using bootstrapping and expectation maximization over clustered documents \cite{mirroshandel2014unsupervised}. Nath et al.\ study cross-document event coreference using semantic transfer and multimodal ensembles \cite{nath2024cdcr}, and a recent survey identifies weak supervision as an important direction for scalable event causality learning \cite{cheng2025survey}. These studies demonstrate the value of indirect supervision, but semantic or temporal correspondence does not by itself establish causality. Therefore, STAGE uses semantic correspondence to construct candidate cross-tweet event pairs and introduces temporal and spatial information separately during relational learning.

\begin{figure*}[t]
    \centering
    \includegraphics[width=\textwidth]{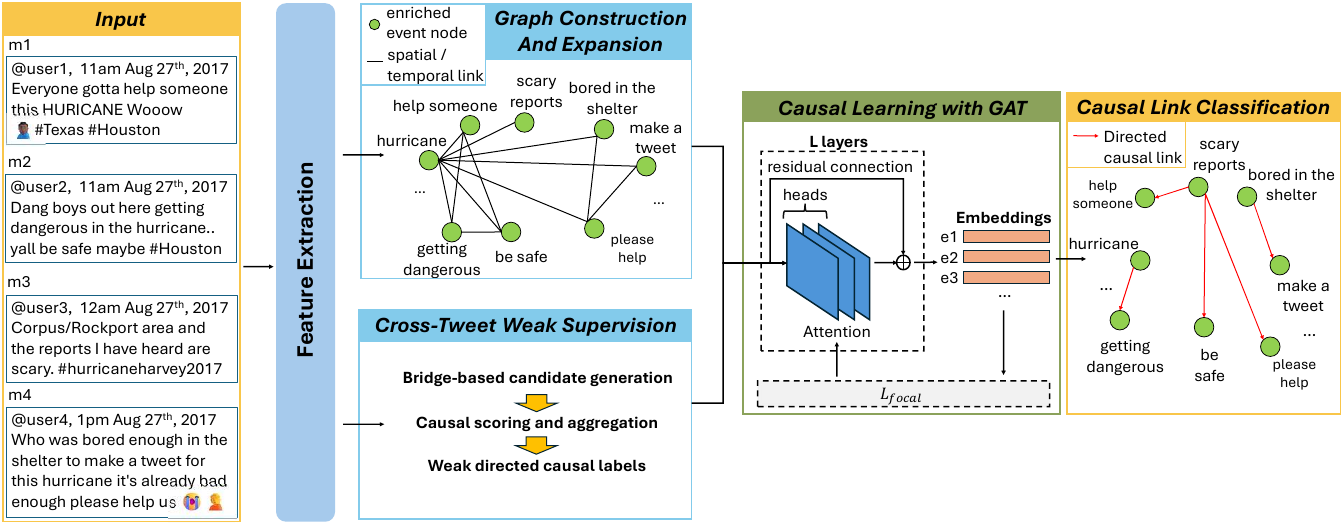}
    \caption{Overview of the STAGE framework. Extracted event representations support both cross-tweet weak supervision, where semantically retrieved candidate pairs are scored and aggregated into weak causal labels with directions assigned to causal pairs, and spatio-temporal graph construction and semantic expansion. A multi-head GAT then learns contextualized event representations for directed causal prediction across tweets.}

    \label{fig:framework}
    \vspace{-2mm}
\end{figure*}

\subsection{Cross-Document and Cross-Tweet Event Linking}

Cross-document event research links event mentions across texts using semantic and structural evidence. Min et al.\ combine local and global reasoning for cross-document event coreference \cite{min-etal-2024-synergetic}, while Xu et al.\ use heterogeneous graph interactions for document-level event extraction \cite{DBLP:journals/corr/abs-2105-14924}. Yao et al.\ introduce argument-centric causal intervention to reduce spurious reliance on lexical similarity in cross-document event coreference \cite{yao2026argument}. Multi-document event extraction has also combined large and small language models to connect heterogeneous documents \cite{min-etal-2025-multi}.

In social-media settings, Dong et al.\ analyze cascading disaster effects using textual and geospatial signals \cite{dong2025}, while Lenti et al.\ model probable causal drivers of climate activism from Reddit and external events \cite{lenti2025}. However, these settings differ from explicit event-occurrence-level causal prediction across independent tweets. Our prior CaST framework studies spatio-temporal event causality within tweets and identifies cross-tweet causal reasoning as an open direction due to the absence of cross-tweet annotations \cite{cast}. STAGE addresses this gap by constructing cross-tweet supervision and learning relations between event occurrences distributed across posts.

\subsection{Knowledge-Augmented and LLM-Based Causal Reasoning}

External knowledge and language models provide additional mechanisms for causal reasoning. Knowledge-graph methods exploit graph structure to infer missing relations \cite{wang2019kgc}, while retrieval-augmented generation introduces retrieved evidence into pretrained language models \cite{lewis2020rag}. Kiciman et al.\ show that LLMs can perform several forms of causal reasoning \cite{kiciman2024}, although evaluation on causal extraction tasks continues to reveal limitations in complex real-world text \cite{luo}.

Structural guidance can improve such reasoning. Long et al.\ use language models as imperfect experts for causal discovery \cite{long2023}, while Roy et al.\ use Causal Event Graphs to support language-based spatio-temporal reasoning \cite{roy}. SERE retrieves structurally relevant demonstrations for LLM-based event causality identification \cite{hao-etal-2026-sere}, and Zhao et al.\ introduce potential-outcome and actual-causality grounding to mitigate causal bias in LLMs \cite{zhao-etal-2026-mitigating}. These approaches demonstrate the value of external structure, but large-scale disaster streams require causal learning over fragmented posts without depending on curated knowledge bases or repeated LLM inference.

\section{Methodology}
\subsection{Problem Formulation}
\label{sub:formulation}

Given a stream of disaster-related tweets $\mathcal{T}=\{t_1,t_2,\ldots\}$, we extract event occurrences $\mathcal{E}_i=\{e_{i1},e_{i2},\ldots\}$ from each tweet $t_i$, where each occurrence denotes a specific event instance through its trigger and argument spans. Let $\mathcal{E}$ denote all extracted event occurrences. Figure~\ref{fig:framework} summarizes the STAGE framework.

The prediction unit is an ordered pair $(e_a,e_b)$. The ordering provides a fixed reference for prediction and does not itself imply causal direction. A pair is cross-tweet when $e_a\in\mathcal{E}_i$ and $e_b\in\mathcal{E}_j$ with $i\neq j$. STAGE predicts causal existence and, for causal pairs, direction:
\begin{equation}
    y^c_{ab}\in\{0,1\}, \qquad
    y^d_{ab}\in\{a\rightarrow b,\; b\rightarrow a\}.
\end{equation}
Thus, the task concerns causal relations between specific event occurrences rather than between tweets as whole units. STAGE separates cross-tweet supervision construction from relational learning. Candidate pairs are constructed from semantic correspondences between event occurrences, while temporal and spatial information are introduced separately through the event graph used for causal learning.

\subsection{Feature Extraction}
\label{sub:feature_extraction}

For each event occurrence $e$, its trigger and argument spans are encoded with CrisisTransformer~\cite{crisis} to obtain the semantic representation $v_e$, which is used for cross-tweet retrieval. Temporal context is encoded from the source-tweet timestamp:
\begin{equation}
\tau_e=
\left[
\frac{\mathrm{hour}}{24},
\frac{\mathrm{weekday}}{6},
\frac{\mathrm{month}}{12},
\frac{\mathrm{day}}{31}
\right].
\end{equation}
Spatial context is represented as:
\begin{equation}
s_e=
\left[
g_e,l_e,c_e,
\frac{\phi_e}{90},
\frac{\lambda_e}{180}
\right],
\end{equation}
where $g_e$ indicates available geolocation metadata, $l_e$ indicates a location identified in the tweet text, $c_e=\min(n_e/5,1)$ is the normalized number of detected location mentions, and $\phi_e$ and $\lambda_e$ are latitude and longitude when explicit numeric coordinates are available; otherwise, their node-feature values are zero. Bounding-box centers, when available, are used separately for spatial graph construction. Note that only $v_e$ is used for cross-tweet semantic retrieval; $\tau_e$ and $s_e$ are introduced during graph construction.

\vspace{-1mm}
\subsection{Cross-Tweet Weak Supervision}
\label{sub:weak_supervision}

Manually annotating cross-tweet causal relations at disaster-stream scale is impractical because the number of possible event-occurrence pairs grows rapidly with the number of extracted events. STAGE therefore constructs cross-tweet training examples from retained within-tweet causal predictions and assigns them weak labels using a fixed causal scorer.

Let $\mathcal{S}$ denote the retained directed local causal predictions from the upstream within-tweet causal pipeline. Each seed $(u,v)\in\mathcal{S}$ represents a predicted relation $u\rightarrow v$, where $u$ is the cause-side event occurrence and $v$ is the effect-side occurrence in the same tweet. Rather than treating the seed itself as cross-tweet supervision, STAGE uses its two endpoints as semantic references for finding related event occurrences elsewhere in the disaster stream.
For each seed endpoint $e\in\{u,v\}$, event representations are L2-normalized and the $k$ most semantically similar occurrences from other tweets are retrieved:
\begin{equation}
\mathcal{N}_k(e)=
\operatorname{TopK}_{o\in\mathcal{E},\,
\mathrm{tweet}(o)\neq\mathrm{tweet}(e)}
\left(\hat v_e^\top \hat v_o\right).
\end{equation}
Here, $\hat v_e^\top\hat v_o$ is cosine similarity because both representations are normalized, and $\mathcal{N}_k(e)$ contains the highest-scoring occurrences outside the source tweet of $e$. Retrieval is performed with FAISS~\cite{faiss} without a fixed similarity threshold. The search spans all extracted events, so a retrieved occurrence may come from a tweet containing no local causal relation.

For a seed $u\rightarrow v$, let $u'\in\mathcal{N}_k(u)$ and $v'\in\mathcal{N}_k(v)$. STAGE constructs three candidate forms:
\begin{equation}
(u,v'), \qquad (u',v), \qquad (u',v').
\end{equation}
The first form replaces the effect-side endpoint of the seed, the second replaces the cause-side endpoint, and the third replaces both. In the two-sided case, $u'$ and $v'$ are paired because they were retrieved relative to the cause and effect endpoints of the same seed; they need not be semantically similar to each other. The two-sided candidate is retained only when
$\mathrm{tweet}(u')\neq\mathrm{tweet}(v')$, ensuring that the final pair is cross-tweet.

\begin{figure}[t]
    \centering
    \includegraphics[width=\columnwidth]{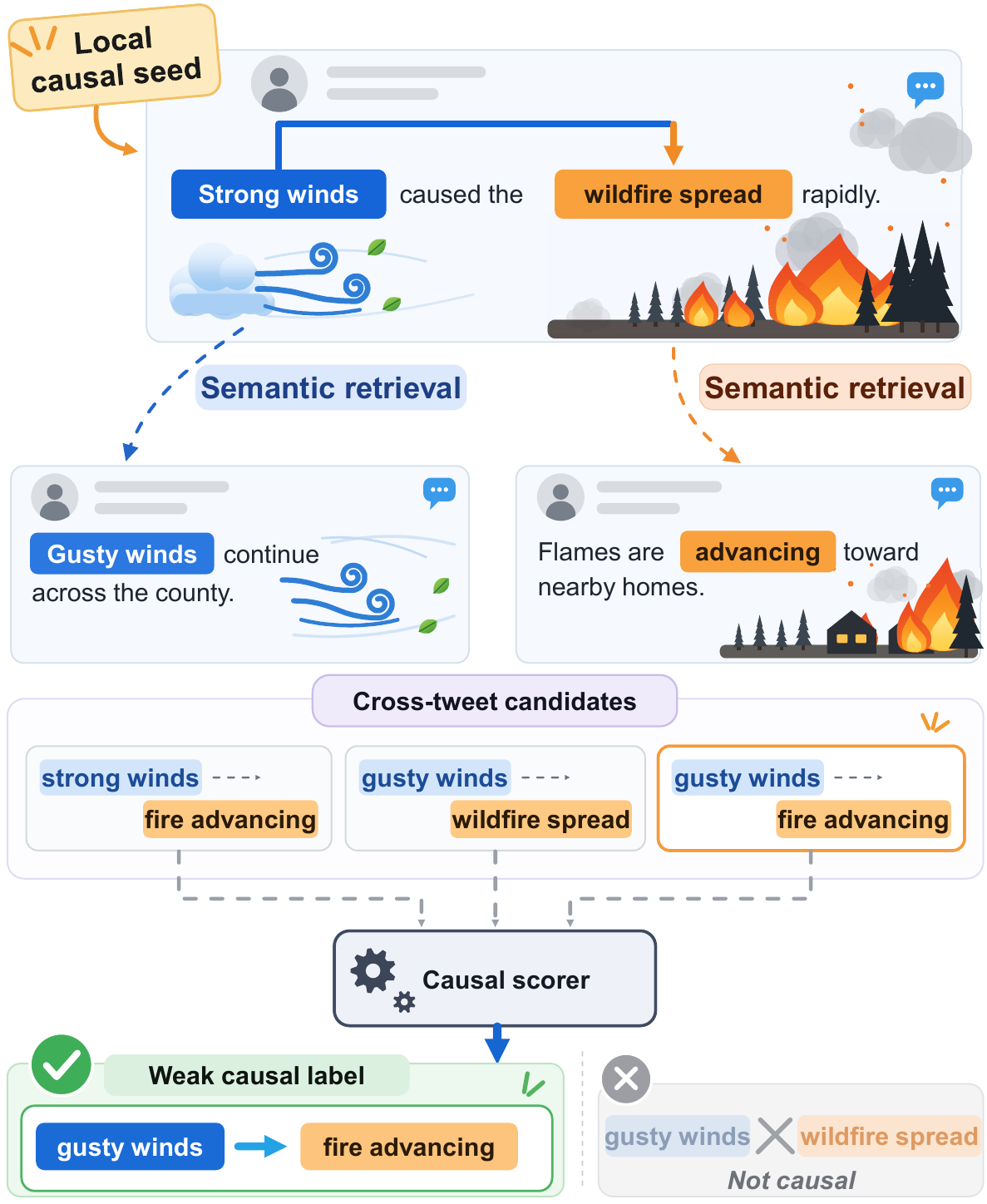}
    \caption{Seed-conditioned cross-tweet weak supervision. Semantic neighbors of local cause-side and effect-side events form cross-tweet candidates, which are evaluated by the causal scorer to produce weak causal labels and directions.}
    \vspace{-3mm}
    \label{fig:weak_supervision}
\end{figure}

The ordering of each candidate follows the cause--effect roles of the originating seed, but this ordering is only a \emph{proposal direction}. Semantic retrieval identifies plausible event correspondences; it does not establish that the resulting pair is causal. Because the Cartesian product of the retrieved neighborhoods can generate many proposals, STAGE retains a bounded set of the highest-ranked candidates for each seed and candidate form using semantic-neighbor rank and similarity support. Figure~\ref{fig:weak_supervision} illustrates this seed-conditioned construction.

Each retained proposal is then evaluated by the fixed cross-tweet causal scorer trained with an external manually annotated resource. This resource is independent of the Wildfire and Hurricane datasets and is treated as an auxiliary component; its construction is outside the focus of this work. The scorer then assigns causal-existence and directional probabilities to the proposal. The same scorer is applied to candidate proposals from both disaster datasets. For a proposal $r$, let $(o_s,o_t)$ denote the ordered candidate endpoints. The proposal representation includes the candidate endpoints, the originating seed endpoints, the candidate form, semantic correspondence information, and proposal-support statistics. Importantly, explicit timestamp, geolocation, temporal distance, spatial distance, and user metadata are not provided to this scorer. For each proposal, it estimates
\begin{equation}
p_r^c=P(y_r^c=1\mid r), \qquad
p_r^d=P(o_s\rightarrow o_t\mid y_r^c=1,r).
\end{equation}
The first quantity, $p_r^c$, measures the probability that a causal relation exists between the two candidate occurrences. The second, $p_r^d$, measures support for the particular direction in which the proposal is presented, conditional on the pair being causal. A final event pair can be generated more than once, for example through different local seeds or different candidate forms. These proposals may also present the same two events in opposite orders. STAGE therefore aggregates proposal-level predictions into a single pair-level prediction.

Let $(e_a,e_b)$ denote a fixed canonical ordering of an unordered event pair, independent of its causal direction, and let $\mathcal{P}_{ab}$ denote all proposals corresponding to that pair. Before aggregation, each proposal-level direction score is aligned to this canonical ordering:
\begin{equation}
\tilde p_r^d=
\begin{cases}
p_r^d, & (o_s,o_t)=(e_a,e_b),\\
1-p_r^d, & (o_s,o_t)=(e_b,e_a).
\end{cases}
\end{equation}
Thus, if a proposal presents the pair in reverse order, its directional probability is flipped so that all proposals express support relative to the same reference ordering.
The aligned proposal scores are then averaged:
\begin{equation}
\bar p_{ab}^c=
\frac{1}{|\mathcal{P}_{ab}|}
\sum_{r\in\mathcal{P}_{ab}}p_r^c,
\qquad
\bar p_{ab}^d=
\frac{1}{|\mathcal{P}_{ab}|}
\sum_{r\in\mathcal{P}_{ab}}\tilde p_r^d.
\end{equation}
Here, $\bar p_{ab}^c$ summarizes causal-existence evidence across all proposals of the pair, while $\bar p_{ab}^d$ summarizes their directional evidence after aligning them to the same ordering. Let $\lambda$ be the causal threshold selected on the scorer development set. The final weak labels are:
\begin{equation}
\hat y_{ab}^c=
\mathbb{1}[\bar p_{ab}^c\ge\lambda],
\qquad
\hat y_{ab}^d=
\begin{cases}
a\rightarrow b, & \bar p_{ab}^d\ge0.5,\\
b\rightarrow a, & \text{otherwise}.
\end{cases}
\end{equation}
The first decision determines whether the event pair is retained as causal or non-causal. The second assigns direction only when the pair is labeled causal.
This process produces a large weakly labeled set of causal and non-causal cross-tweet event pairs. Importantly, weak-supervision construction remains separate from STAGE's explicit spatio-temporal modeling: semantic retrieval determines which pairs are considered, the fixed causal scorer determines their weak labels, and temporal and spatial information are introduced later through the event graph. Human-gold Wildfire and Hurricane evaluation pairs are excluded from weak training; Human--Human reliability and Weak--Gold agreement are reported in Section~\ref{subsec:dataset}.

\subsection{Spatio-Temporal Graph Construction and Semantic Expansion}
\label{sub:graph_construction}

STAGE updates event representations as new disaster information becomes available and constructs a contextual graph for cross-tweet learning. Event occurrences are the graph nodes; semantic information is incorporated into node representations, while temporal and spatial proximity define the message-passing structure.

\subsubsection{Semantic Expansion}

To incorporate semantic context from earlier observations, STAGE maintains a memory $\mathcal{M}_t$ of previously observed event occurrences. For an event $e$ with semantic representation $v_e$, let $\mathcal{R}_q(e)\subseteq\mathcal{M}_t$ denote its $q$ most semantically similar previously observed events. Their representations are aggregated as
\begin{equation}
z_e=
\frac{1}{|\mathcal{R}_q(e)|}
\sum_{m\in\mathcal{R}_q(e)} v_m.
\end{equation}
Here, $z_e$ summarizes semantic context from the retrieved events and is set to zero when no semantic neighbor is available. The event representation is then expanded as
\begin{equation}
\tilde v_e = v_e + z_e.
\end{equation}
Thus, $\tilde v_e$ combines the event's own representation with information from semantically related earlier observations. After processing $e$, the event is added to $\mathcal{M}_t$ for use by subsequent events. The expanded semantic representation is combined with the temporal and spatial features from Section~\ref{sub:feature_extraction}:
\begin{equation}
\tilde x_e = \tilde v_e \,\Vert\, \tau_e \,\Vert\, s_e ,
\end{equation}
where $\Vert$ denotes concatenation. Semantic expansion determines the information carried by each node, while the graph relations below determine which event occurrences exchange contextual information.

\subsubsection{Spatio-Temporal Context Graph}

The chronologically ordered event stream is divided into overlapping windows. For each window $w$, let $V_w\subseteq\mathcal{E}$ denote its event occurrences and define
\begin{equation}
G_w=(V_w,E_w), \qquad
E_w=E_w^T\cup E_w^S.
\end{equation}
Here, $E_w^T$ and $E_w^S$ denote temporal and spatial event-to-event relations. Two events $e_i,e_j\in V_w$ are temporally connected when $|t(e_i)-t(e_j)|\leq\Delta_t$ and spatially connected when $d(e_i,e_j)\leq\Delta_s$, where $t(e)$ is the source-tweet timestamp and $d(\cdot,\cdot)$ is geographic distance. These edges encode contextual proximity rather than causal direction; events without coordinates simply receive no spatial edges. Weakly supervised causal pairs are used only as prediction targets and are not added as message-passing edges, keeping the contextual graph separate from causal supervision.

\subsection{Causal Learning with GAT}
\label{sub:causal_learning}

Given the contextual graph $G_w$ and node representations $\tilde{x}_e$ from Section~\ref{sub:graph_construction}, STAGE applies a multi-head Graph Attention Network (GAT)~\cite{gat} over the temporal and spatial relations:
\begin{equation}
h_e=\operatorname{GAT}^{(L)}(\tilde{x}_e,G_w),
\end{equation}
where $L$ denotes the number of GAT layers. The resulting representation $h_e$ combines the event's node features with contextual information propagated from neighboring events.

STAGE then evaluates causal relations at the event-pair level. For an ordered pair $(e_a,e_b)$, the contextualized representations are concatenated and passed to a feed-forward scorer:
\begin{equation}
p_{ab}=\sigma\!\left(f_\theta(h_a\Vert h_b)\right),
\end{equation}
where $p_{ab}$ represents support for the directed relation $e_a\rightarrow e_b$. Because the scorer is order-sensitive, applying it to the reversed pair yields the corresponding support for $e_b\rightarrow e_a$. Causal existence is evaluated using a threshold selected on the development set, while direction is determined from the relative support assigned to the two orderings. The model is trained on ordered event pairs using binary focal loss~\cite{focalloss}:
\begin{equation}
\mathcal{L}_{\mathrm{focal}}
=
-\alpha_t(1-p_t)^\gamma \log p_t,
\end{equation}
where $p_t=p_{ab}$ for a positive ordered relation and $p_t=1-p_{ab}$ otherwise, while $\alpha_t=\alpha$ for positive examples and $\alpha_t=1-\alpha$ for negative examples. The focusing parameter $\gamma$ down-weights easy examples. For a causal pair, the weakly supervised direction determines which ordering is treated as positive, while the reverse ordering is treated as negative. For a non-causal pair, both orderings are treated as negative. Direction is therefore learned through the asymmetric treatment of the two orderings rather than through a separate direction head.

\section{Experiments}
\subsection{Dataset}
\label{subsec:dataset}

\paragraph{Dataset Collection and Statistics}
We evaluate STAGE on two disaster-related social media corpora: California Wildfire (Wildfire) and Hurricane Harvey (Hurricane). Wildfire spans November 7 to December 1, 2018, whereas Hurricane spans August 26 to August 31, 2017. Although the two corpora contain comparable amounts of social-media data, Hurricane is substantially more temporally concentrated. Table~\ref{tab:dataset_statistics} summarizes the main dataset and weak-supervision statistics.

\begin{table}[t]
\centering
\caption{Summary statistics for the Wildfire and Hurricane datasets.}
\label{tab:dataset_statistics}
\resizebox{0.9\columnwidth}{!}{
\begin{tabular}{lrr}
\toprule
Statistic & Wildfire & Hurricane \\
\midrule
Tweets & 186,365 & 202,908 \\
Event occurrences & 173,159 & 236,833 \\
UTC calendar dates & 25 & 6 \\
Events per calendar day & 6,926 & 39,472 \\
Local causal seeds & 49,368 & 56,963 \\
Cross-tweet candidate pairs & 4,401,845 & 4,831,069 \\
Weak-supervision pool & 1,328,610 & 1,519,001 \\
Weak causal rate (\%) & 36.85 & 28.75 \\
Human-gold pairs & 9,826 & 9,407 \\
\bottomrule
\end{tabular}}
\vspace{-3mm}
\end{table}

The retained local causal predictions serve as seeds for the cross-tweet weak-supervision construction described in Section~\ref{sub:weak_supervision}. Seed-conditioned semantic retrieval generates cross-tweet candidate pairs, which are subsequently evaluated by the causal scorer and aggregated into the weak-supervision pools. Candidate and weak-supervision counts in Table~\ref{tab:dataset_statistics} refer to unique event pairs. The human-gold evaluation sets are kept separate from the weak supervision used to train STAGE. The weak-supervision pools are described in Section~\ref{subsec:experimental_setup}.

\paragraph{Human-Gold Annotation}
To construct the human-gold evaluation sets, we sample 10,000 event pairs from the full cross-tweet candidate pool of each disaster and manually annotate them. Each pair is assigned one of five substantive labels: \textsc{A-Causes-B}, \textsc{B-Causes-A}, \textsc{Related-Noncausal}, \textsc{Same-Event}, or \textsc{Unrelated}; pairs for which a reliable judgment cannot be made are marked \textsc{Uncertain}. For binary causal evaluation, \textsc{A-Causes-B} and \textsc{B-Causes-A} are mapped to the causal class, while \textsc{Related-Noncausal}, \textsc{Same-Event}, and \textsc{Unrelated} are mapped to the non-causal class. \textsc{Uncertain} cases are excluded from evaluation. The resulting gold sets contain 3,609 causal Wildfire pairs (36.73\%) and 3,964 causal Hurricane pairs (42.14\%); their total usable sizes are reported in Table~\ref{tab:dataset_statistics}. All human-gold pairs are removed from the candidate pool before constructing the weak-supervision training and development sets and are used only for final evaluation. Consequently, the evaluation measures causal prediction within the retrieved candidate space and does not quantify causal relations missed by candidate generation.

To assess annotation reliability, 2,000 pairs from the human-gold annotation pools (1,000 from each disaster) are independently annotated by two annotators. This subset is used only to assess human--human annotation reliability and does not contribute to model training or weak-label construction. After excluding 40 \textsc{Uncertain} cases, causal-existence agreement is evaluated on the remaining pairs, while direction is evaluated only on pairs independently identified as causal by both annotators. The resulting agreement and Cohen's $\kappa$ scores are reported in Table~\ref{tab:annotation_agreement}.

\paragraph{Weak-Supervision Validation}
We assess weak-supervision quality by comparing the weak labels with the final human-gold labels on both datasets. Table~\ref{tab:annotation_agreement} reports both human--human annotation reliability (H--H) and weak-label agreement with human gold (W--Gold). For direction, agreement is evaluated only on pairs identified as causal by both sources being compared.
\begin{table}[t]
\centering
\caption{Human annotation (H) reliability and agreement between weak supervision (W) and human-gold labels (Gold).}
\label{tab:annotation_agreement}
\resizebox{\columnwidth}{!}{
\begin{tabular}{lllccc}
\toprule
Comparison & Dataset & Task & $N$ & Agreement (\%) & $\kappa$ \\
\midrule
H--H & -- & Causal existence & 1,960 & 98.88 & 0.976 \\
H--H & -- & Direction & 725 & 99.31 & 0.986 \\
\midrule
W--Gold & Wildfire & Causal existence & 9,826 & 73.60 & 0.490 \\
W--Gold & Wildfire & Direction & 3,287 & 96.20 & 0.904 \\
W--Gold & Hurricane & Causal existence & 9,407 & 65.91 & 0.255 \\
W--Gold & Hurricane & Direction & 1,470 & 99.66 & 0.992 \\
\bottomrule
\end{tabular}}
\vspace{-3mm}
\end{table}
Additionally, weak labels show stronger agreement with human gold for causal existence on Wildfire than on Hurricane, indicating that weak-label quality varies across disasters. In contrast, when both weak supervision and human annotation identify a pair as causal, their directional judgments show consistently high agreement. This suggests that the main disagreement lies in determining whether a causal relation exists, rather than in its direction once causality is identified.

\subsection{Baselines}
\label{subsec:baselines}
We compare STAGE with baselines from three model families:
\begin{itemize}
    \item \textit{Statistical and neural baselines}: ANM~\cite{anm}, RF, SVM, BiLSTM+Attention~\cite{bilstmatt}, and BERT~\cite{bert}.
    \item \textit{Graph-, prompt-, and knowledge-based baselines}: DECI~\cite{gcn}, PPAT~\cite{ppat}, DAPrompt~\cite{daprompt}, CHEER~\cite{cheer}, RCE~\cite{rce}, and KnowQA~\cite{knowqa}.
    \item \textit{Event-causality-specific baselines}: SemDI~\cite{semdi}, ICCL~\cite{iccl}, iLIF~\cite{ilif}, and GLM4ECI~\cite{glm4eci}.
\end{itemize}
For the weak cross-tweet benchmark, STAGE and all supervised baselines are trained on the same constructed weak-supervision pairs and evaluated on the same human-gold sets, so the comparison isolates differences in downstream learning rather than available supervision. Non-parametric or unsupervised methods use their original inference procedures.
\newcolumntype{C}{>{\centering\arraybackslash}X}
\begin{table*}[t]
\centering
\caption{Causal discovery under in-domain weak cross-tweet supervision.
\textbf{Bold}: best; \underline{underline}: second best.}
\label{tab:main}

\begin{tabularx}{\dimexpr\textwidth-2\tabcolsep\relax}
{@{}l *{6}{C} p{5pt} *{6}{C}@{}}
\toprule
& \multicolumn{6}{c}{\textbf{Wildfire}}
&& \multicolumn{6}{c}{\textbf{Hurricane}} \\
\cmidrule(lr){2-7}
\cmidrule(lr){9-14}

Method
& AUPRC & AUROC & P & R & F1 & DirAcc
&& AUPRC & AUROC & P & R & F1 & DirAcc \\
\midrule

\textit{All-positive reference}
& 0.367 & 0.500 & 0.367 & 1.000 & 0.537 & --
&& 0.421 & 0.500 & 0.421 & 1.000 & 0.593 & -- \\

\midrule

ANM
& 0.364 & 0.496 & 0.367 & 1.000 & 0.537 & 0.420
&& 0.419 & 0.497 & 0.421 & 0.998 & 0.592 & 0.440 \\

RF
& 0.666 & 0.769 & 0.546 & 0.753 & 0.633 & 0.920
&& 0.595 & 0.686 & 0.574 & 0.574 & 0.574 & 0.953 \\

SVM
& 0.659 & 0.766 & 0.581 & 0.701 & 0.636 & 0.942
&& 0.590 & 0.684 & 0.575 & 0.536 & 0.555 & 0.955 \\

BiLSTM+Attn
& \textbf{0.836}
& \underline{0.878}
& \textbf{0.816}
& 0.699
& \textbf{0.753}
& \textbf{0.999}
&& 0.616
& 0.709
& 0.658
& 0.386
& 0.486
& 0.995 \\

BERT
& 0.725
& 0.804
& 0.672
& 0.643
& 0.657
& 0.989
&& \textbf{0.626}
& \textbf{0.723}
& \underline{0.661}
& 0.450
& 0.536
& \textbf{1.000} \\

\midrule

DECI
& 0.369 & 0.504 & 0.369 & 0.985 & 0.536 & 0.969
&& 0.418 & 0.494 & 0.421 & 1.000 & 0.593 & 0.975 \\

PPAT
& 0.375 & 0.515 & 0.377 & 0.757 & 0.503 & 0.798
&& 0.478 & 0.595 & 0.481 & 0.642 & 0.550 & 0.905 \\

DAPrompt
& 0.751 & 0.828 & 0.672 & 0.695 & 0.683 & 0.970
&& 0.589 & 0.685 & 0.587 & 0.545 & 0.565 & 0.976 \\

CHEER
& 0.375 & 0.481 & 0.377 & 0.932 & 0.537 & 0.915
&& 0.454 & 0.536 & 0.432 & 1.000 & 0.603 & 0.931 \\

RCE
& 0.649 & 0.740 & 0.691 & 0.577 & 0.629 & 0.965
&& 0.519 & 0.582 & 0.537 & 0.396 & 0.456 & 0.965 \\

KnowQA
& 0.663
& 0.765
& 0.512
& 0.787
& 0.621
& 0.981
&& 0.598
& 0.702
& 0.563
& 0.687
& \textbf{0.618}
& 0.991 \\

\midrule

SemDI
& 0.432 & 0.598 & 0.417 & 0.814 & 0.552 & 0.520
&& 0.450 & 0.546 & 0.430 & 0.881 & 0.577 & 0.477 \\

ICCL
& 0.610
& 0.744
& 0.572
& 0.696
& 0.628
& 0.983
&& 0.556
& 0.670
& 0.526
& 0.727
& \underline{0.610}
& 0.967 \\

iLIF
& 0.678 & 0.761 & 0.684 & 0.586 & 0.631 & 0.986
&& 0.566 & 0.667 & 0.546 & 0.636 & 0.587 & 0.981 \\

GLM4ECI
& 0.577 & 0.724 & 0.487 & 0.814 & 0.610 & 0.576
&& 0.442 & 0.528 & 0.422 & 0.981 & 0.590 & 0.656 \\

\midrule
\textbf{STAGE (ours)}
& \underline{0.828}
& \textbf{0.885}
& \underline{0.769}
& 0.737
& \underline{0.752}
& \underline{0.991}
&& \underline{0.620}
& \underline{0.710}
& \textbf{0.676}
& 0.333
& 0.446
& \underline{0.996} \\

\bottomrule
\end{tabularx}
\vspace{-1.5mm}
\end{table*}

\vspace{-3mm}
\subsection{Evaluation Metrics}
\label{subsec:evaluation_metrics}

We evaluate causal-existence prediction using Precision, Recall, F1, area under the precision--recall curve (AUPRC), and area under the receiver operating characteristic curve (AUROC). We treat AUPRC as the primary threshold-independent metric because causal pairs form the minority class and the precision--recall curve directly reflects the trade-off between recovering causal relations and introducing false causal links. AUROC is reported as a complementary measure of overall class separability. Precision, Recall, and F1 characterize performance at the causal decision threshold selected on the development set and fixed before evaluation on the human-gold test set. We additionally report Direction Accuracy (DirAcc) among human-gold causal pairs that are predicted as causal. DirAcc is the proportion of these detected causal pairs for which the predicted cause--effect direction matches the human annotation, separating direction selection from causal-existence recall. Recall is not ranked independently, since high recall can result from over-predicting the causal class.

\subsection{Experimental Setup}
\label{subsec:experimental_setup}

After removing all human-gold pairs, we construct event-disjoint training and development splits from the remaining weak-supervision pool so that an event occurrence cannot contribute supervision to both partitions; a pair is retained only when both endpoints belong to the same partition. Wildfire contains 454,369 training and 8,776 development pairs, with causal rates of 32.02\% and 30.52\%, respectively. Hurricane contains 404,351 training and 7,730 development pairs, with causal rates of 19.64\% and 19.00\%.
We use \texttt{crisistransformers/CT-M2-Complete} as the semantic encoder with maximum input length 192 and encoding batch size 128. Semantic expansion retrieves the top $q=3$ related events. Temporal and spatial graph connections are restricted to 12 hours and 150\,km, respectively, with at most eight temporal and eight spatial neighbors per event. Two-hop neighborhoods are sampled with fanouts $[20,10]$.
The GAT uses hidden dimension 256, eight attention heads, and dropout 0.1. Training uses batch size 128 and AdamW with learning rate $5\times10^{-3}$, weight decay $10^{-5}$, and focal-loss focusing parameter $\gamma=2$. All models are trained for 15 epochs using multiple random seeds, and we report the mean performance across runs. For each run, the checkpoint with highest development-set AUPRC is retained. Experiments are conducted on an Intel Core i9-13950HX processor, 64\,GB RAM, and an NVIDIA RTX 5000 Ada Generation Laptop GPU with 16\,GB memory.

\section{Results}\label{sec:results}

\begin{figure*}[t]
\centering
\includegraphics[width=0.9\textwidth]{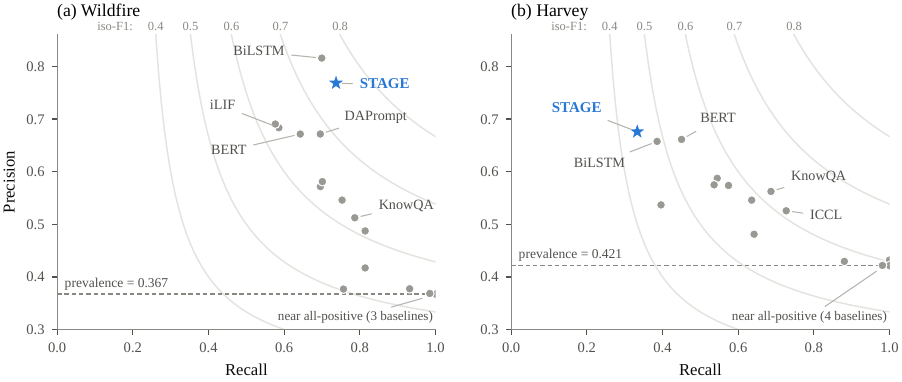}
\caption{Precision--recall operating points under weak cross-tweet supervision on (a) Wildfire and (b) Hurricane. Gray curves are iso-F1 contours; the dashed line is the causal-class prevalence, which is also the precision of a predictor that labels every pair as causal.}
\label{fig:pr}
\end{figure*}

We report causal-existence in three settings: training on the weak cross-tweet supervision of a disaster and testing on that disaster's human-gold pairs (Table~\ref{tab:main}); training only on local within-tweet causal supervision and testing on cross-tweet pairs; and cross-disaster transfer (Table~\ref{tab:generalization}). Direction is additionally evaluated in the in-domain weak cross-tweet benchmark. Component ablations are reported in Table~\ref{tab:ablation}. Because causal pairs form 36.73\% of Wildfire and 42.14\% of Hurricane gold pairs, an all-positive predictor already achieves F1 scores of 0.537 and 0.593, respectively. We therefore emphasize AUPRC, with AUROC as a complementary ranking metric, and interpret F1 together with the precision--recall operating point in Fig.~\ref{fig:pr}.

\begin{table*}[t]
\centering
\caption{Generalization beyond in-domain weak cross-tweet supervision.
\emph{Local $\rightarrow$ cross-tweet}: within-tweet training, same-disaster cross-tweet evaluation.
\emph{Cross-disaster}: weak cross-tweet training on one disaster, human-gold evaluation on the other.}
\label{tab:generalization}
\setlength{\tabcolsep}{3.6pt}
\resizebox{0.97\textwidth}{!}{%
\begin{tabular}{l ccc c ccc c ccc c ccc}
\toprule
& \multicolumn{7}{c}{\textbf{Local $\rightarrow$ cross-tweet}} && \multicolumn{7}{c}{\textbf{Cross-disaster transfer}} \\
\cmidrule(lr){2-8}\cmidrule(lr){10-16}
& \multicolumn{3}{c}{Wildfire} && \multicolumn{3}{c}{Hurricane} && \multicolumn{3}{c}{Hurricane $\rightarrow$ Wildfire} && \multicolumn{3}{c}{Wildfire $\rightarrow$ Hurricane} \\
\cmidrule(lr){2-4}\cmidrule(lr){6-8}\cmidrule(lr){10-12}\cmidrule(lr){14-16}
Method & AUPRC & AUROC & F1 && AUPRC & AUROC & F1 && AUPRC & AUROC & F1 && AUPRC & AUROC & F1 \\
\midrule
\textit{All-positive reference} & 0.367 & 0.500 & 0.537 &  & 0.421 & 0.500 & 0.593 &  & 0.367 & 0.500 & 0.537 &  & 0.421 & 0.500 & 0.593 \\
\midrule
ANM & 0.364 & 0.496 & 0.527 &  & 0.419 & 0.497 & 0.590 &  & 0.364 & 0.496 & 0.537 &  & 0.419 & 0.497 & 0.593 \\
RF & \textbf{0.552} & 0.663 & 0.552 &  & 0.542 & 0.609 & 0.566 &  & 0.547 & 0.700 & 0.328 &  & 0.527 & 0.641 & 0.486 \\
SVM & 0.456 & 0.575 & 0.539 &  & 0.484 & 0.557 & 0.547 &  & 0.566 & 0.711 & 0.361 &  & 0.533 & 0.636 & 0.365 \\
BiLSTM+Attn & 0.509 & \underline{0.693} & 0.583 &  & 0.540 & \underline{0.652} & 0.611 &  & \textbf{0.634} & 0.729 & 0.427 &  & \underline{0.562} & 0.654 & 0.209 \\
BERT & 0.504 & 0.669 & 0.564 &  & \underline{0.544} & 0.632 & 0.591 &  & 0.603 & \underline{0.746} & 0.377 &  & 0.508 & 0.596 & 0.151 \\
\midrule
DECI & 0.365 & 0.495 & 0.537 &  & 0.416 & 0.491 & 0.593 &  & 0.363 & 0.491 & 0.537 &  & 0.422 & 0.503 & 0.590 \\
PPAT & 0.451 & 0.563 & 0.537 &  & 0.391 & 0.456 & 0.589 &  & 0.485 & 0.654 & 0.570 &  & 0.430 & 0.516 & 0.533 \\
DAPrompt & 0.455 & 0.658 & 0.580 &  & 0.540 & 0.638 & 0.588 &  & \underline{0.615} & \textbf{0.748} & 0.552 &  & 0.537 & 0.660 & 0.119 \\
CHEER & 0.389 & 0.515 & 0.551 &  & 0.447 & 0.529 & 0.603 &  & 0.405 & 0.545 & 0.549 &  & 0.436 & 0.506 & 0.589 \\
RCE & 0.464 & 0.609 & 0.528 &  & 0.493 & 0.557 & 0.538 &  & 0.516 & 0.646 & 0.295 &  & 0.487 & 0.576 & 0.038 \\
KnowQA & 0.523 & 0.679 & 0.584 &  & \textbf{0.587} & \textbf{0.697} & 0.633 &  & 0.578 & 0.737 & 0.597 &  & 0.535 & 0.647 & 0.584 \\
\midrule
SemDI & 0.386 & 0.553 & 0.539 &  & 0.424 & 0.512 & 0.595 &  & 0.444 & 0.594 & 0.538 &  & 0.457 & 0.556 & 0.584 \\
ICCL & \underline{0.538} & \textbf{0.702} & 0.580 &  & 0.539 & 0.638 & 0.591 &  & 0.360 & 0.491 & 0.537 &  & 0.559 & \underline{0.668} & 0.511 \\
iLIF & 0.486 & 0.669 & 0.583 &  & 0.535 & 0.635 & 0.552 &  & 0.502 & 0.691 & 0.598 &  & 0.472 & 0.562 & 0.058 \\
GLM4ECI & 0.361 & 0.487 & 0.537 &  & 0.422 & 0.501 & 0.593 &  & 0.404 & 0.547 & 0.537 &  & 0.498 & 0.607 & 0.539 \\
\midrule
\textbf{STAGE (ours)} & 0.488 & 0.615 & 0.531 &  & 0.513 & 0.602 & 0.592 &  & 0.500 & 0.702 & 0.422 &  & \textbf{0.582} & \textbf{0.675} & 0.466 \\
\bottomrule
\end{tabular}}
\vspace{-2mm}
\end{table*}

\subsection{Weak Cross-Tweet Supervision}
\paragraph{California Wildfire} STAGE achieves the highest AUROC (0.885) and second-highest AUPRC (0.828), while its F1 is effectively tied with BiLSTM+Attention (0.752 versus 0.753). This indicates strong causal/non-causal separation, although STAGE does not uniformly outperform the strongest text-only baseline. Several baselines perform near chance in the fragmented cross-tweet setting, including ANM, DECI, PPAT, CHEER, and SemDI (AUROC 0.481--0.598). These methods label 72\%--100\% of test pairs as causal, which drives recall toward 1 but reduces precision toward the causal-class prevalence in Fig.~\ref{fig:pr}a, keeping their F1 (at most 0.552) close to the all-positive reference of 0.537.

\paragraph{Hurricane Harvey} STAGE and the strongest baselines show lower causal-ranking performance on Harvey, consistent with its denser event stream, stronger class imbalance, and noisier weak labels. STAGE ranks second in both AUPRC (0.620) and AUROC (0.710), behind BERT (0.626 and 0.723) and marginally ahead of BiLSTM+Attention (0.616 and 0.709). STAGE has the highest precision of all methods (0.676) but also the lowest recall (0.333), which yields an F1 of 0.446, below the all-positive reference. The low F1 reflects the operating threshold more than ranking quality. The decision threshold $\lambda$ is selected on weak-labeled development data, and the causal rate of Harvey's weak training pairs (19.64\%) is less than half that of the human-gold pairs (42.14\%). STAGE labels 20.8\% of Hurricane test pairs as causal, close to the training prior, so its errors are predominantly missed causal pairs rather than spurious causal edges (Fig.~\ref{fig:pr}b).

\vspace{-2mm}
\paragraph{Direction} Conditional on detecting a causal pair, STAGE predicts its direction correctly in 99.1\% of Wildfire cases and 99.6\% of Hurricane cases. This is consistent with the high directional agreement between weak and human-gold labels in Table~\ref{tab:annotation_agreement}, suggesting that most remaining difficulty lies in determining whether a causal relation exists rather than in orienting an identified causal pair.

\subsection{Cross-Tweet Versus Local Supervision}

Trained only on local within-tweet causal supervision, STAGE generalizes poorly to cross-tweet pairs (Table~\ref{tab:generalization}, left). Its AUPRC is 0.488 on Wildfire and 0.513 on Hurricane (sixth and eighth of 16), and it labels most cross-tweet pairs as causal (recall 0.945 and 0.933). The strongest results in this setting come from text-only models: RF (0.552 AUPRC) and ICCL (0.702 AUROC) on Wildfire, and KnowQA (0.587 AUPRC, 0.697 AUROC) on Harvey. Local supervision contains no labeled causal relation whose endpoints occur in different tweets, so STAGE receives no direct supervision for cross-tweet causal decisions.
\begin{figure}[t]
    \centering
    \includegraphics[width=\columnwidth]{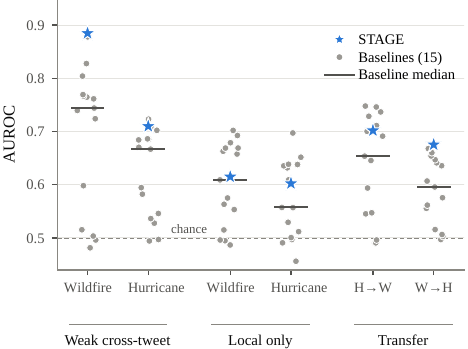}
    \caption{AUROC of STAGE and baselines across the six evaluation settings.}
    \vspace{-5mm}
    \label{fig:settings}
\end{figure}
Replacing local supervision with the constructed cross-tweet weak supervision reverses this picture (Fig.~\ref{fig:settings}). STAGE improves by $+0.340$ AUPRC and $+0.270$ AUROC on Wildfire and by $+0.107$ AUPRC and $+0.108$ AUROC on Harvey, the largest AUPRC gain of any method on both datasets. The supervision also benefits the baselines, with median AUPRC gains of $+0.14$ on Wildfire and $+0.03$ on Harvey. These results indicate that the constructed cross-tweet supervision provides useful learning signal independently of the downstream architecture, while STAGE can additionally exploit that signal through relational modeling.

\vspace{-1mm}
\subsection{Cross-Disaster Transfer}

Transferring from Wildfire to Harvey, STAGE achieves the best AUPRC (0.582) and AUROC (0.675) of all methods, ahead of BiLSTM+Attention (0.562 AUPRC) and ICCL (0.668 AUROC). It loses only 0.038 AUPRC relative to in-domain training on Harvey, whereas BERT, which has the highest in-domain Hurricane AUPRC and AUROC, loses 0.118 (0.626 to 0.508). Several baselines collapse toward a single class under this shift: BERT, DAPrompt, iLIF, and RCE recover fewer than 9\% of causal pairs, while ANM, DECI, and CHEER label nearly all pairs as causal.

The reverse direction is substantially harder for STAGE: Hurricane-to-Wildfire transfer yields AUPRC 0.500 and AUROC 0.702, with AUPRC falling by 0.328 relative to in-domain Wildfire training, while several baselines transfer more effectively. Hurricane has a denser event stream, a more class-imbalanced weak-training split, and lower Weak--Gold agreement ($\kappa=0.255$ versus $0.490$), while STAGE also operates conservatively after transfer (recall $0.329$). Because several baselines trained on the same supervision transfer better, the asymmetry may also reflect STAGE's dataset-specific context features. Its temporal and spatial encodings contain absolute calendar fields and coordinates, which shift substantially across disasters occurring in different months and regions. Relative temporal and spatial representations may therefore improve cross-disaster transfer.

\definecolor{deltaNeg}{RGB}{180,45,45}
\definecolor{deltaPos}{RGB}{35,125,75}

\newcommand{\drop}[1]{\,{\scriptsize\bfseries\textcolor{deltaNeg}{($-$#1)}}}
\newcommand{\gain}[1]{\,{\scriptsize\bfseries\textcolor{deltaPos}{($+$#1)}}}

\begin{table}[t]
\centering
\caption{Ablation of STAGE under weak cross-tweet supervision.
Parentheses give the change relative to the full model.}
\label{tab:ablation}
\setlength{\tabcolsep}{3pt}
\small
\begin{tabular}{l ccc}
\toprule
Variant & AUPRC & AUROC & F1 \\
\midrule

\multicolumn{4}{l}{\textit{Wildfire}} \\
\hspace{0.6em}\textbf{STAGE (full)}
& 0.828
& 0.885
& 0.752 \\

\hspace{0.6em}w/o sem. expansion
& 0.803\drop{0.025}
& 0.865\drop{0.020}
& 0.735\drop{0.017} \\

\hspace{0.6em}w/o spatial context
& 0.807\drop{0.021}
& 0.870\drop{0.015}
& 0.738\drop{0.014} \\

\hspace{0.6em}w/o temporal context
& 0.811\drop{0.017}
& 0.872\drop{0.013}
& 0.743\drop{0.009} \\

\hspace{0.6em}Pair-only (w/o GAT)
& 0.819\drop{0.009}
& 0.880\drop{0.005}
& 0.751\drop{0.001} \\

\midrule

\multicolumn{4}{l}{\textit{Harvey}} \\
\hspace{0.6em}\textbf{STAGE (full)}
& 0.620
& 0.710
& 0.446 \\

\hspace{0.6em}w/o sem. expansion
& 0.616\drop{0.004}
& 0.703\drop{0.007}
& 0.427\drop{0.019} \\

\hspace{0.6em}w/o spatial context
& 0.618\drop{0.002}
& 0.704\drop{0.006}
& 0.488\gain{0.042} \\

\hspace{0.6em}w/o temporal context
& 0.617\drop{0.003}
& 0.704\drop{0.006}
& 0.450\gain{0.004} \\

\hspace{0.6em}Pair-only (w/o GAT)
& 0.606\drop{0.014}
& 0.689\drop{0.021}
& 0.477\gain{0.031} \\

\bottomrule
\end{tabular}
\vspace{-4mm}
\end{table}

\subsection{Ablation Study}

The ablations reveal different component contributions across the two disasters (Table~\ref{tab:ablation}). On Wildfire, removing semantic expansion causes the largest degradation ($-0.025$ AUPRC, $-0.020$ AUROC), while removing GAT message passing has a smaller effect ($-0.009$, $-0.005$). On Hurricane, the pattern reverses: removing GAT produces the largest degradation ($-0.014$ AUPRC, $-0.021$ AUROC), while other removals change AUPRC by at most 0.004. Overall, the results suggest that semantic enrichment and relational message passing contribute differently across disasters, while the full model achieves the highest mean AUPRC and AUROC.

\section{Conclusion}

This paper introduced STAGE, a framework for cross-tweet causal discovery that combines scalable weak-supervision construction with spatio-temporal relational learning. The results show that the constructed cross-tweet supervision substantially improves over local-only supervision and also benefits supervised baselines, while STAGE provides one effective relational instantiation for learning from this supervision. The results also reveal limitations: weak-label quality varies across disasters, and cross-disaster transfer remains sensitive to dataset shift and absolute spatio-temporal representations. Overall, STAGE provides a scalable approach to learning causal relations across fragmented disaster reports. Future work will focus on more transferable spatio-temporal representations and on calibrating cross-disaster weak supervision to shifts in causal prevalence and weak-label reliability.

\section*{Acknowledgment}
This work was partially supported by the U.S. National Science Foundation (NSF) under Grants No.~2219614 and No.~2624573.

\bibliographystyle{IEEEtran}
\bibliography{references}

\end{document}